%% file: dexsuite.tex
\pdfoutput=1
\documentclass[letterpaper, 10 pt, conference]{ieeeconf}
\IEEEoverridecommandlockouts
\usepackage{amsmath,amsfonts,bm}
\usepackage{graphicx}
\usepackage{array}
\usepackage{booktabs}
\usepackage{multirow}
\usepackage{makecell}
\usepackage{pifont}
\usepackage{xcolor}
\usepackage{textcomp}
\usepackage{stfloats}
\usepackage{url}
\usepackage{cite}
\usepackage{longtable}

\usepackage{float}
\usepackage[bookmarks=true]{hyperref}

\definecolor{sdgray}{gray}{0.5}
\newcolumntype{L}[1]{>{\raggedright\arraybackslash}p{#1}}
\newcommand{\cmark}{\ding{51}}
\newcommand{\xmark}{\ding{55}}

\begin{document}

\title{\LARGE \bf DexSuite: An All-in-One Framework for Dexterous Manipulation Across Arms, Hands, and Tasks}

\author{Anas Houssaini$^{1,2}$, Hugo He$^{2}$, Junming Shi$^{1,2}$, Amin Abyaneh$^{1,2}$, Ricardo Chahine$^{2}$,\\
Shuo Wen$^{1,2}$, Mohamad H. Danesh$^{1,2}$, Charles-\'Etienne Fr\'echette$^{1,2}$, Mariana Sosa Guzm\'an$^{1,2}$,\\
Mathias Desrochers$^{2}$, Rayan Houssaini$^{1}$, Adam Landa$^{1}$, Th\'eophile Souli\'e$^{2}$,\\
Charlotte Morissette$^{1,2}$, Annie Tims$^{1,2}$, Martino Russi$^{3}$, Jonathan Lussier$^{4}$,\\
Gregory Dudek$^{1,2}$, Doina Precup$^{1,2}$, Julen Urain$^{5}$, Hsiu-Chin Lin$^{1,2}$, and David Meger$^{1,2}$%
\thanks{$^{1}$Mila; $^{2}$McGill University; $^{3}$Hugging Face; $^{4}$Kinova; $^{5}$Amazon FAR. E-mail: achraf.elhoussaini@mail.mcgill.ca.}}

\makeatletter
\let\@IEEEmaketitleorig\@maketitle
\renewcommand{\@maketitle}{%
  \@IEEEmaketitleorig
  \vspace{0.3\baselineskip}
  \noindent\includegraphics[width=\textwidth]{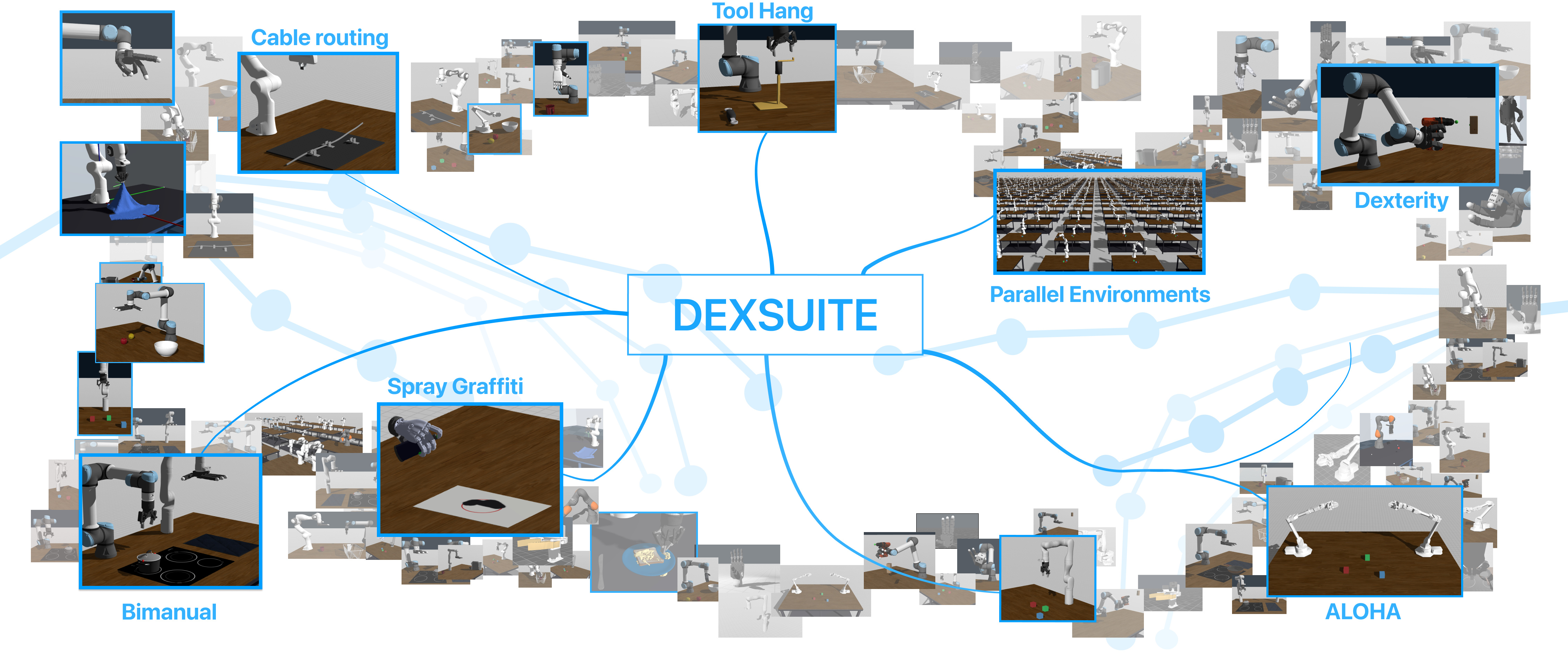}\par
  \setcounter{figure}{0}%
  \refstepcounter{figure}%
  \label{fig:concept}%
  \@makecaption{\fnum@figure}{%
    \textbf{DexSuite}. An all-in-one simulation framework for dexterous manipulation,
    supporting diverse embodiments, GPU-parallel simulation, and
    cross-embodiment training and evaluation. It includes a benchmark of 16 task-embodiment pairs, a dataset of 4{,}700 successful episodes, and training and evaluation pipelines.%
  }\par
  \vspace{-0.9\baselineskip}%
}
\makeatother

\maketitle
\thispagestyle{empty}
\pagestyle{empty}
\bstctlcite{IEEEexample:BSTcontrol}

\begin{abstract}
	Human-level dexterity remains a central open problem in robotics, yet few simulation frameworks cover the full workflow of dexterous manipulation, from setting up a robot to collecting dexterous demonstrations and evaluating policies. We present DexSuite, an all-in-one simulation framework for dexterous manipulation built on the Genesis simulator. DexSuite provides 30 tasks and assembles its 10 arms and 12 end-effectors, from parallel-jaw grippers to dexterous hands, into more than 100 single-arm robots and thousands of bimanual pairs. It also ships a standardized environment API, integrated teleoperation, and simulated tactile sensing. Moreover, we define a benchmark of 16 task-embodiment pairs spanning diverse contact complexities and horizons, and release a dataset of 4{,}700 successful episodes in simulation from human teleoperation, scripted experts, and learned policies. We release DexSuite as open source and hope others will build on it. Videos and datasets are available at \url{https://dexsuiteorg.github.io}.

\end{abstract}


\section{Introduction}

As humans, we designed the physical world for dexterous hands \cite{bicchi2000hands,okamura2000overview}. Parallel-jaw grippers have enabled impressive progress in pick-and-place style manipulation, but they remain limited when tasks require in-hand adjustment, multi-point stabilization, or tool-like interactions \cite{billard2019trends,intelligence2025pi_}. If robots are to operate in existing environments rather than in environments reshaped around their limitations, dexterous hands must be treated as a first-class benchmark target.

\input{Dexsuite/comparaison_table}

Recent advances in simulation, perception, and learning have accelerated progress in robotic manipulation \cite{makoviychuk2021isaac,mittal2025isaac,gu2023maniskill2,tao2024maniskill3,zhu2020robosuite,james2020rlbench}. Since large-scale hardware experimentation remains slow and failure-prone, simulation is the most practical path to scalable training and controlled evaluation. Nevertheless, most widely used simulation benchmarks continue to prioritize parallel-jaw grippers \cite{zhu2020robosuite,james2020rlbench,gu2023maniskill2,nasiriany2024robocasa,yu2020meta,liu2023libero,mees2022calvin,mandlekar2023mimicgen,mu2025robotwin,li2023behavior}, as do the largest teleoperation datasets \cite{mandlekar2021matters,khazatsky2024droid,o2024open,walke2023bridgedata,fang2024rh20t,brohan2022rt}. In part, this is because dexterous hands are harder to teleoperate, train, and evaluate given their high-dimensional actuation, reliance on intermittent contacts, and sensitivity to modeling errors \cite{andrychowicz2020learning,rajeswaran2017learning,bao2023dexart}. Controlled studies across embodiments therefore remain difficult. Multi-finger benchmarks are often specialized for a specific hand model on a fixed set of tasks \cite{li2023dexdeform}, end-effectors are rarely swapped under consistent success metrics, and existing research in bimanual dexterity is spread across different frameworks \cite{grotz2024peract2,chen2022towards}.

We present DexSuite (Fig.~\ref{fig:concept}), which standardizes task semantics and includes a benchmark with teleoperated demonstrations to support reproducible learning and comparison across embodiments. We use the benchmark to study (i) whether policies trained for a dexterous hand get the same success rates as policies trained for a gripper on identical tasks, (ii) to what extent long-horizon and sequential-contact tasks reduce hand success more than short tasks, and finally (iii) which task types relying on finger-level control remain challenging for standard imitation learning. Our \textbf{contributions} are threefold.

\noindent \textbf{DexSuite framework.} Built on Genesis \cite{Genesis}, a lightweight and Python-native simulator, DexSuite provides 30 tasks, a standardized environment API, integrated teleoperation, and simulated tactile sensing.

\noindent \textbf{Benchmark.} A curated benchmark of 16 task-embodiment pairs designed to evaluate dexterous manipulation across action dimensionality, temporal horizon, and contact-rich control, with baselines for five imitation learning methods.

\noindent \textbf{DexSuite dataset.} A release of 4{,}700 successful episodes from human teleoperation, scripted experts and learned policies.

\section{Related Work}

\subsection{Manipulation simulators}

Isaac Lab \cite{mittal2025isaac} and ManiSkill3 \cite{tao2024maniskill3} enable GPU-parallel simulation for scalable robot learning. The robosuite \cite{zhu2020robosuite} and RLBench \cite{james2020rlbench} frameworks provide large task families and standardized APIs.  RoboCasa \cite{nasiriany2024robocasa} and OmniGibson \cite{li2023behavior} supply high-quality and realistic scenes. However, dexterous-hand and arm-with-gripper tasks are typically separated, and bimanual benchmarks are fragmented across suites. While some suites include both dexterous hands and grippers, swapping end-effectors while keeping the same scene and conditions is uncommon. When several configurations are supported, hand-versus-gripper comparison on identical tasks is usually not the primary benchmark interface \cite{zhu2020robosuite}.  
Table~\ref{tab:framework-comparison} summarizes the capabilities of major simulation frameworks.

\subsection{Dexterous manipulation benchmarks}

DexArt \cite{bao2023dexart} benchmarks arm-mounted hands on articulated objects, DexMV converts third-person human videos into robot-hand demonstrations \cite{qin2022dexmv}, and Dactyl \cite{andrychowicz2020learning} and DAPG \cite{rajeswaran2017learning} established in-hand reorientation with reinforcement learning and demonstration-augmented reinforcement learning from teleoperated data, respectively. These efforts each evaluate a single hand model. For bimanual dexterity, Bi-DexHands provides two-hand tasks with free-floating Shadow Hands on Isaac Gym \cite{makoviychuk2021isaac,chen2022towards}, while PerAct2 \cite{grotz2024peract2,james2020rlbench} and robosuite \cite{zhu2020robosuite} provide dual-arm tabletop tasks with parallel-jaw grippers. Concurrent work such as LabDex \cite{tang2026labdex} benchmarks dexterous manipulation in laboratory settings, in both simulation and the real world.

\subsection{Datasets for imitation and offline learning}

Large-scale datasets such as DROID \cite{khazatsky2024droid} and Open X-Embodiment \cite{o2024open} have demonstrated the impact of teleoperation data on learning generalist policies, but they use parallel-jaw grippers, focus on arm-level behaviors \cite{mandlekar2021matters}, and are not designed to compare grippers and dexterous hands on identical tasks. Hand-centric datasets such as DexDeform \cite{li2023dexdeform} and DexMV \cite{qin2022dexmv} are tied to specific embodiments and task families, while large synthetic grasp datasets \cite{mahler2017dex,fang2020graspnet,wang2022dexgraspnet,ye2025dex1b} provide static grasps or short-horizon motions rather than full manipulation trajectories under a consistent observation and action interface.

DexSuite is designed to fill this gap. It integrates GPU-parallel simulation, rigid, articulated and deformable dynamics, simulated tactile sensing, native dexterous and bimanual setups, and on-the-fly modular robot composition within a single stack. DexSuite defines embodiment-independent tasks and success conditions. This allows the integration of a large number of robot configurations for every task. We pair this simulation framework with a fixed benchmark protocol containing full-episode teleoperation demonstrations for dexterous hands.

\section{DexSuite}
\label{sec:dexsuite}

\begin{figure*}[!t]
    \centering
    \includegraphics[width=\textwidth]{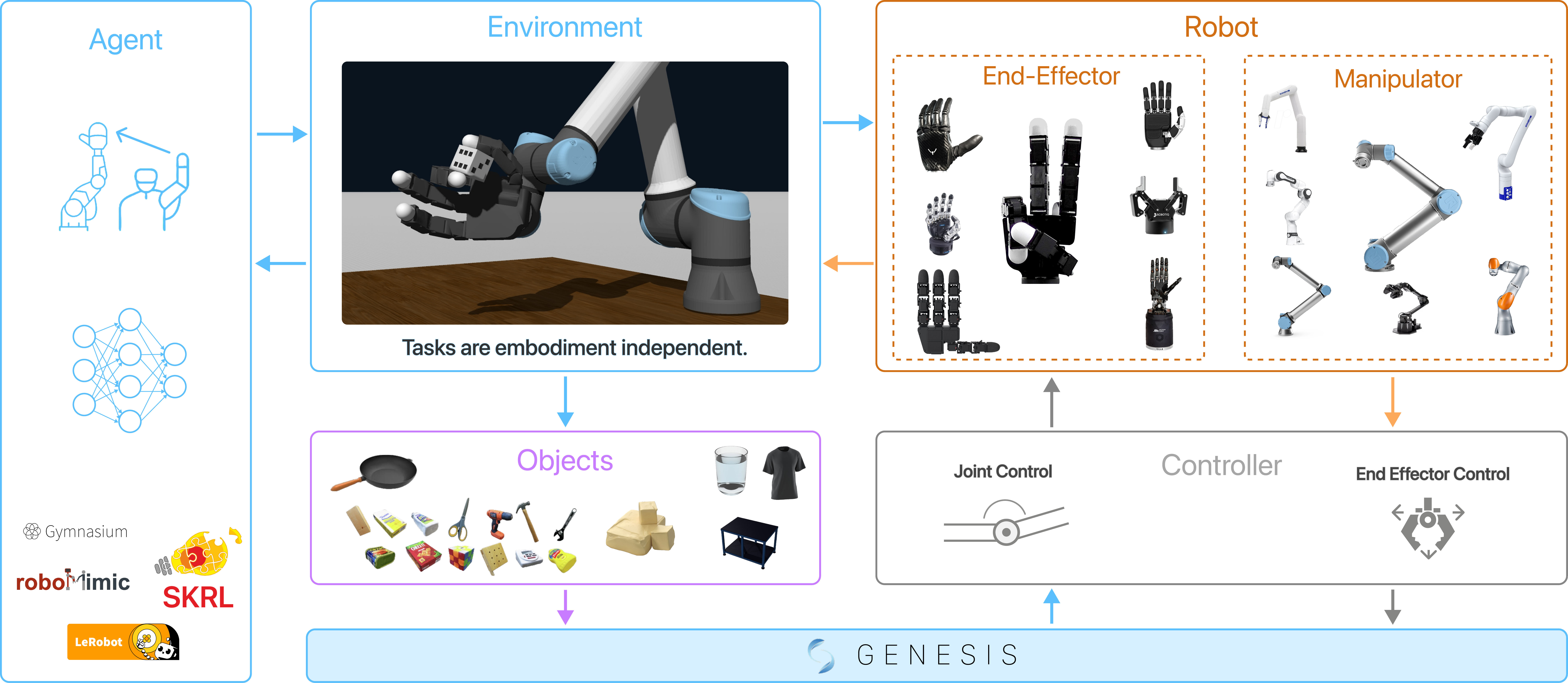}
    \caption{DexSuite architecture. Tasks define assets, reset logic, observations, actions, and success metrics. Robot embodiments and controllers are instantiated through a shared abstraction on top of Genesis.}
    \label{fig:overview}
    \vspace{-4mm}
\end{figure*}

\begin{figure}[!b]
    \centering
    \includegraphics[width=\linewidth]{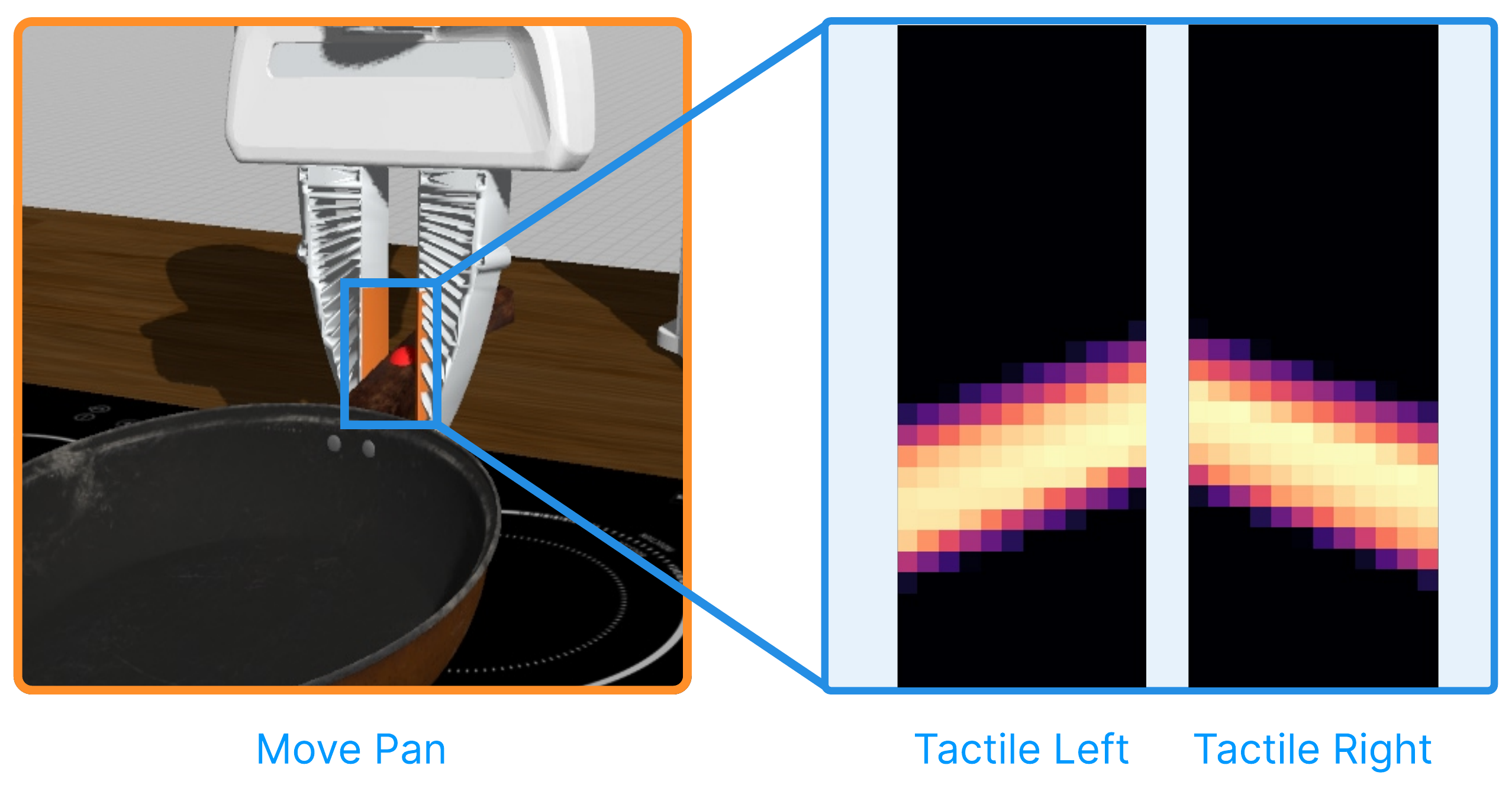}
    \caption{Simulated readings from the two FlexiTac pads of the UMI-style gripper (Right) during the pan task (Left).}
    \label{fig:tactile}
\end{figure}

DexSuite is a simulation framework built on top of Genesis \cite{Genesis}. Genesis provides the underlying physics, rendering, asset loading, and parallel simulation, while DexSuite provides the manipulation layer needed to turn these primitives into reproducible robot learning experiments. As shown in Fig.~\ref{fig:overview}, this layer includes a library of tasks and robot embodiments, embodiment-independent task definitions, modular arm and end-effector composition, workspace-aware initialization, tactile sensing (Fig.~\ref{fig:tactile}), teleoperation, and a unified Gymnasium API. DexSuite further provides a benchmark and curated dataset of dexterous manipulation demonstrations for training and evaluation.

\subsection{Environments}
  DexSuite includes a total of 30 tasks designed to evaluate various aspects of robotic manipulation. These are categorized into two physical domains requiring different control strategies.

\begin{itemize} 
\item \textbf{Rigid objects (23 tasks).} Fundamental skills such as reaching, lifting, pushing, pick-and-place, stacking, and object reorientation. These tasks emphasize precision grasping and contact stability.
      
\item \textbf{Articulated objects (7 tasks).} Interactions with mechanisms that have internal degrees of freedom, such as folding glasses, opening a jar, pressing a toaster lever or closing a window sash. These require an understanding of kinematic constraints and applying forces along specific trajectories.\end{itemize}

We also integrate the YCB object and model set \cite{calli2015ycb}, add a set of new objects with tuned masses and centers of mass, and provide textures with varying friction and light reflection properties.

\subsection{Embodiment Abstraction}

Robot manipulation frameworks often implicitly hard-code task logic and object initialization to a specific robot kinematic structure \cite{tao2024maniskill3,mittal2025isaac}. DexSuite addresses this by adapting environment constraints to the loaded robot, rather than imposing static workspace assumptions. The framework provides 10 arms alongside integrated robots and a library of end-effectors, most of them adapted from MuJoCo Menagerie~\cite{zakka2022mujoco}. Modular composition supports single-arm and bimanual configurations for both existing and new tasks.

\begin{figure*}[!t]
    \centering
    \includegraphics[width=\textwidth]{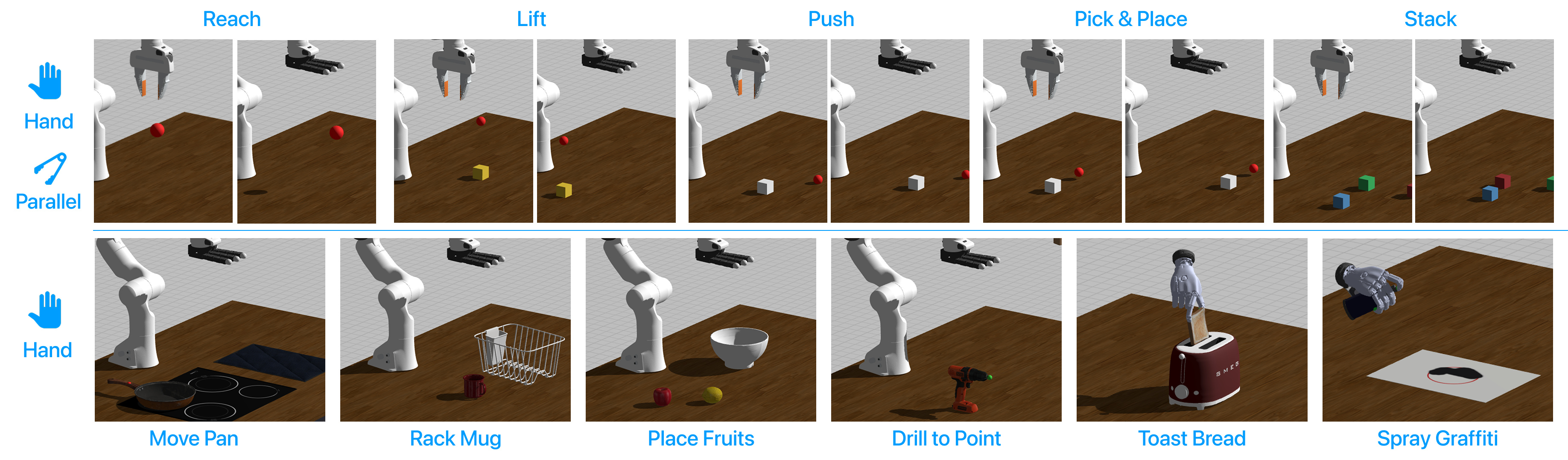}
    \caption{Tasks contained in the DexSuite benchmark. The five at the top are for cross-embodiment evaluation; the bottom ones target multi-fingered dexterous capability.}
    \label{fig:benchmark_tasks}
\end{figure*}

\noindent\textbf{Robot composition.} DexSuite abstracts robot construction into modular and integrated definitions to support a variety of asset formats. Independent arm and end-effector models are combined to create modular robots at runtime. Regardless of the underlying robot definition, this abstraction offers a common control interface that enables the arm to be controlled via inverse kinematics and the hand through joint-position commands. The same idea applies to task design where DexSuite exposes a single Tool Center Point (TCP) for all end-effectors, which is the midpoint of the fingertip for grippers or a central reachable frame for dexterous hands, in order to define rewards and success conditions regardless of embodiment (Fig.~\ref{fig:benchmark_tasks}).

\noindent\textbf{Workspace-aware sampling.} Many frameworks hard-code object sampling regions to a specific robot workspace, causing invalid initializations when changing embodiments. DexSuite instead uses Workspace-Aware Sampling, assigning each arm a reachable Workspace Axis-Aligned Bounding Box (AABB). This allows task initialization to automatically adapt across embodiments while remaining kinematically feasible.

\subsection{Data Collection and Interfaces}

\begin{figure}[!b]
    \centering
    \includegraphics[width=0.85\linewidth]{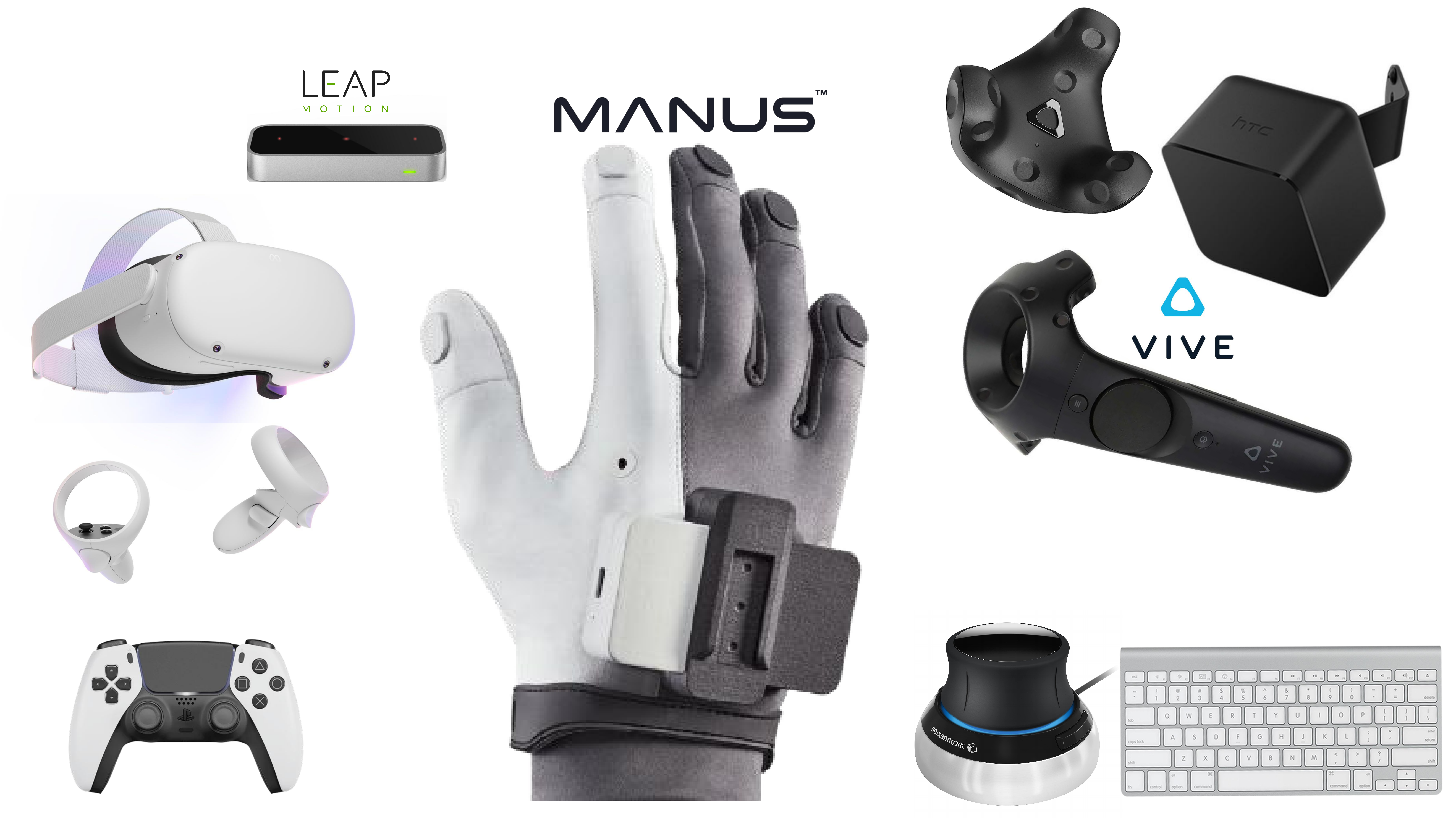}
    \caption{Teleoperation devices integrated in DexSuite.}
    \label{fig:teleop_devices}
\end{figure}

\noindent\textbf{Teleoperation.} DexSuite supports teleoperation for both arm and finger control (Fig.~\ref{fig:teleop_devices}). Arm pose can be controlled with a SpaceMouse, HTC Vive, or Meta Quest controllers, while finger motion can be captured using MediaPipe \cite{lugaresi2019mediapipe}, Leap Motion, Meta Quest hand tracking, or Manus MetaGlove Pro gloves. Related systems such as BiDex \cite{shaw2024bimanual} and TeleDex \cite{rayyan2026teledex} teleoperate dexterous hands with motion-capture gloves and smartphones.

\noindent\textbf{Tactile sensing.} DexSuite simulates the FlexiTac tactile sensor~\cite{huang2026flexitac} (Fig.~\ref{fig:tactile}). A tactile pad is a grid of taxels on a flat elastomer, 12 by 32 at 2\,mm spacing by default as on the real pad, and each taxel converts its penetration into nearby objects into a normal and a friction force. Pads can be declared on any end-effector link with a flat contact surface, with configurable grid size and spacing. We ship tactile pads on a UMI-style parallel-jaw gripper~\cite{chi2024universal} and on a tactile ALOHA manipulator, and the scripted gripper demonstrations record both pads at every step. The Sharpa demonstrations record per-finger contact forces instead. 

\noindent\textbf{Extensibility.} DexSuite allows task, robot, controller, sensor, and simulation configuration through a single \texttt{make} interface. New embodiments can be composed from existing arms and end-effectors without modifying task logic, allowing platforms such as bimanual ALOHA \cite{zhao2023learning} and DROID \cite{khazatsky2024droid} to be instantiated within the same framework. All environments follow the Gymnasium interface \cite{towers2026gymnasium}, enabling direct integration with existing learning pipelines.

\section{Benchmark \& Dataset}
\subsection{Benchmark Design} 
\label{sec:bench_design}

To address the three questions raised in the introduction, namely the embodiment gap, robustness to long horizons, and contact-rich capability, the benchmark contains 11 tasks organized into two evaluation tracks (Fig.~\ref{fig:benchmark_tasks}). Five tasks are simple enough to be attempted with either end-effector and are completed with both a parallel-jaw gripper and the Allegro hand. The other six are completed with a dexterous hand only, because their geometry precludes the use of a simple gripper. Four of them use the 16-DoF Allegro hand and the final two use the 22-DoF Sharpa Wave hand due to their extreme geometric complexity. This gives 16 task-embodiment pairs.

\textbf{Track 1. Cross-Embodiment Evaluation.} The five shared tasks use identical task definitions and success criteria for both end-effectors.
    
\textbf{Track 2. Multi-fingered Dexterous Capability.} The six hand-only tasks target stable grasping, constrained transport, tool use, multi-object interaction, and long-horizon manipulation. 

Together, the two tracks separate embodiment effects from challenges that specifically require dexterous hands, while spanning tasks of increasing contact complexity and temporal horizon.

\subsection{The DexSuite Dataset}
\label{sec:dexsuite_dataset}

\begin{figure}[!b]
    \centering
    \includegraphics[width=0.85\linewidth]{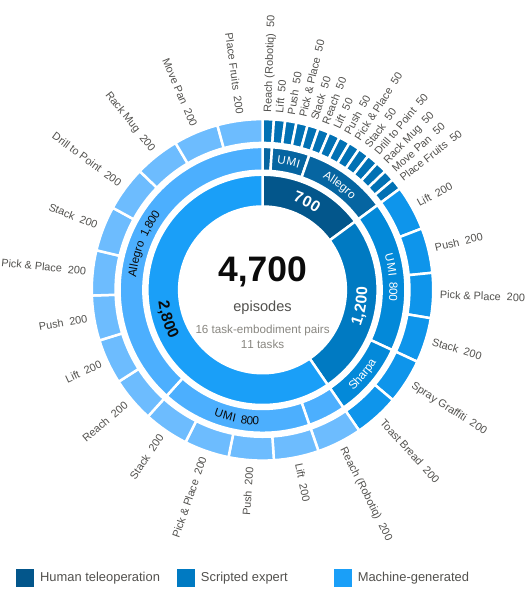}
    \caption{Composition of the DexSuite dataset by collection tier, embodiment, and task. Each human task-embodiment pair contains 50 episodes and each scripted or machine-generated pair contains 200. All episodes are recorded at 20\,Hz with front and wrist cameras.}
    \label{fig:dataset}
    \vspace{-2mm}
\end{figure}

We release 4{,}700 successful episodes covering the 16 task-embodiment pairs of the benchmark (Fig.~\ref{fig:dataset}), collected across three tiers: human teleoperation, scripted experts, and learned-policy rollouts. Together, these tiers support a range of data regimes within a single benchmark. The human tier comprises 700 demonstrations on the Robotiq, UMI, and Allegro embodiments. The scripted tier adds 1{,}200 demonstrations using the UMI gripper and Sharpa hand. The machine-generated tier contributes 2{,}800 successful policy rollouts.

All trajectories share a common observation and action format containing proprioception, object states, and front- and wrist-camera observations. Scripted tactile configurations additionally include fingertip tactile readings. The dataset is released in LeRobot format \cite{cadene2026lerobot}, as well as HDF5 format compatible with robomimic \cite{mandlekar2021matters}.

\section{Experiments and Results}

\input{Dexsuite/imitation_results}

\subsection{Experiment Setup}
\label{sec:results_il}

We evaluate imitation learning baselines on the benchmark introduced in Sec. \ref{sec:bench_design}. We use the robomimic \cite{mandlekar2021matters} implementations of behavior cloning \cite{pomerleau1988alvinn} with Gaussian (BC-G), Gaussian mixture (BC-GMM), and Transformer (BC-T) policies, alongside Diffusion Policy \cite{chi2025diffusion} and ACT \cite{zhao2023learning}. Models are trained on 50 human demonstrations for all tasks except the last two long-horizon dexterous tasks, where we use 200 scripted demonstrations per configuration, as described in Sec.~\ref{sec:dexsuite_dataset}. The BC baselines use state observations, ACT uses images and proprioception, and diffusion is evaluated with both modalities. During training, each run is rolled out for 50 episodes every 50 epochs and we keep the checkpoint with the highest rollout success, as in robomimic~\cite{mandlekar2021matters}. Unlike robomimic, which reports this maximum, we evaluate the selected checkpoint again on three new evaluation seeds with 50 episodes each and report the mean success rate and standard deviation. For ACT, the in-training maximum exceeds the success of the same checkpoint on the new seeds by 7 percentage points on average across the 14 pairs trained on human demonstrations, and the last checkpoint was the best one in only 2 of them. With 50 episodes, the standard error of a success rate can reach 7 points, and the 150 evaluation episodes bring it to about 4, so gaps of a few points between policies should be read with care. The spread across seeds is similar for both embodiments, 4.8 and 5.3 points for the gripper and the Allegro hand on Track 1 without \textit{Stack}, and we recommend the same protocol for work using the benchmark.

\subsection{Results}
\label{sec:answers}

\paragraph{\textbf{Embodiment gap}}
The hand's higher action dimensionality does not consistently reduce success (Table~\ref{tab:il_results}). All six policy--observation combinations favor Allegro on \textit{Push}, while four favor the gripper on \textit{Lift}. State-based performance on \textit{Reach} differs by at most four percentage points across embodiments. On \textit{Pick \& Place}, the gap changes direction with the policy: BC-T performs better with the hand (76\% versus 64\%), while ACT performs better with the gripper (66\% versus 60\%). The embodiment gap therefore depends on both the task and the policy.

\begin{figure}[!h]
    \centering
    \includegraphics[width=\linewidth]{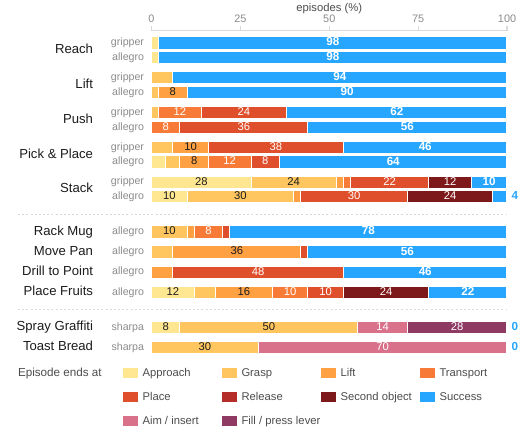}
    \caption{Stage-wise outcomes of low-dimensional diffusion policies over 50 diagnostic rollouts per configuration, separate from the evaluations in Table~\ref{tab:il_results}. Bars show the percentage of episodes ending at each stage, with successes in blue. The gripper is the Robotiq for \textit{Reach} and the UMI gripper otherwise. For Sharpa, aim and fill refer to \textit{Spray Graffiti}, and insert and press lever refer to \textit{Toast Bread}.}
    \label{fig:stages}
\end{figure}

\paragraph{\textbf{Horizon and sequential manipulation}}
Performance drops sharply on tasks requiring successive object interactions. The best state-based policies reach only 16\% with the gripper and 9\% with Allegro on \textit{Stack}, while image policies reach at most 2\%. On \textit{Place Fruits}, the best state-based result is 9\%, compared with 27\% for ACT. Yet ACT reaches 89\% on \textit{Move Pan}, despite its comparable horizon of 500 steps versus 600 for \textit{Stack}. This suggests that coordinating successive interactions, rather than duration alone, contributes to the difficulty.

The diagnostic rollouts in Fig.~\ref{fig:stages}, scored by the stages of each task, show failures during these transitions. On \textit{Stack}, 22--30\% of episodes fail at the first placement and 12--24\% at the second object across embodiments. On \textit{Place Fruits}, 24\% fail on the second fruit. Policies therefore struggle to carry an initial manipulation sequence through to subsequent objects.

\paragraph{\textbf{Dexterous capability}}
Several hand-only tasks are learned successfully: state-based diffusion reaches 82\% on \textit{Rack Mug}, and ACT reaches 89\% on \textit{Move Pan}. Precise tool use remains harder. State-based diffusion reaches 50\% on \textit{Drill to Point}, with 48\% of diagnostic rollouts failing at final alignment. On \textit{Spray Graffiti}, half of the diagnostic rollouts fail at grasping the can, and another 28\% paint without reaching the required coverage. No baseline exceeds 1\% on \textit{Spray Graffiti}, and all score 0\% on \textit{Toast Bread}.

\section{Conclusion} \label{sec:conclusion}

We introduced DexSuite, a framework built on Genesis, together with a benchmark of 16 task-embodiment pairs and a dataset of 4{,}700 successful episodes. Policies trained and evaluated on a dexterous hand match or outperform their gripper counterparts on several shared tasks, while precise tool use and sequential object manipulation remain difficult. To enable repeatable evaluation across tasks and embodiments, we make the framework, benchmark, and dataset available as open source.

Contributors can add new environments as one class on the standard API, and new arms, hands, and sensors without changing any task code, then test them on the standard tasks. The more groups build on DexSuite, the more comparable results in dexterous manipulation become, and every new task, robot, or method makes it more useful to everyone.

\bibliographystyle{IEEEtran}
\bibliography{references}

\end{document}

%% file: Dexsuite/comparaison_table.tex
\begin{table*}[!t]
    \centering
    \caption{Comparison of manipulation frameworks and integrated resources. }
    \label{tab:framework-comparison}
    \footnotesize
    \setlength{\tabcolsep}{3pt}
    \renewcommand\arraystretch{1.03}
    
    \begin{tabular}{@{}L{2.2cm} L{1.6cm} c c c c c c c c c@{}}
    \toprule
    \textbf{Name} & \textbf{Physics} &
    \textbf{Parallel} &
    \makecell{\textbf{Rigid}\\\textbf{Deform/Fluid}} &
    \makecell{\textbf{Robot}\\\textbf{Library}} &
    \makecell{\textbf{Dex}\\\textbf{Hands}} &
    \textbf{Bimanual} &
    \makecell{\textbf{Embodiment}\\\textbf{Independent}} &
    \makecell{\textbf{Hand}\\\textbf{Dataset}} &
    \makecell{\textbf{Hand}\\\textbf{Teleop}} &
    \makecell{\textbf{Tactile}\\\textbf{Sensing}} \\
    \midrule

    \textbf{DexSuite} & Genesis &
    {\color{green}\cmark} &
    {\color{green}\cmark} &
    {\color{green}\cmark} &
    {\color{green}\cmark} &
    {\color{green}\cmark} &
    {\color{green}\cmark} &
    {\color{green}\cmark} &
    {\color{green}\cmark} &
    {\color{green}\cmark} \\
    
    \midrule
    robosuite & MuJoCo &
    {\color{red}\xmark} &
    {\color{red}\xmark} &
    {\color{green}\cmark} &
    {\color{green}\cmark} &
    {\color{green}\cmark} &
    {\color{green}\cmark} &
    {\color{red}\xmark} &
    {\color{red}\xmark} &
    {\color{red}\xmark} \\

    Isaac Lab & PhysX &
    {\color{green}\cmark} &
    {\color{red}\xmark} &
    {\color{green}\cmark} &
    {\color{green}\cmark} &
    {\color{green}\cmark} &
    {\color{red}\xmark} &
    {\color{red}\xmark} &
    {\color{green}\cmark} &
    {\color{green}\cmark} \\

    RLBench & CoppeliaSim &
    {\color{red}\xmark} &
    {\color{red}\xmark} &
    {\color{green}\cmark} &
    {\color{red}\xmark} &
    {\color{red}\xmark} &
    {\color{red}\xmark} &
    {\color{red}\xmark} &
    {\color{red}\xmark} &
    {\color{red}\xmark} \\

    ManiSkill3 & SAPIEN/\allowbreak PhysX &
    {\color{green}\cmark} &
    {\color{red}\xmark} &
    {\color{green}\cmark} &
    {\color{green}\cmark} &
    {\color{green}\cmark} &
    {\color{red}\xmark} &
    {\color{red}\xmark} &
    {\color{red}\xmark} &
    {\color{green}\cmark} \\

    OmniGibson & PhysX &
    {\color{green}\cmark} &
    {\color{green}\cmark} &
    {\color{green}\cmark} &
    {\color{green}\cmark} &
    {\color{green}\cmark} &
    {\color{green}\cmark} &
    {\color{red}\xmark} &
    {\color{green}\cmark} &
    {\color{red}\xmark} \\

    RoboCasa & MuJoCo &
    {\color{red}\xmark} &
    {\color{red}\xmark} &
    {\color{green}\cmark} &
    {\color{red}\xmark} &
    {\color{red}\xmark} &
    {\color{red}\xmark} &
    {\color{red}\xmark} &
    {\color{red}\xmark} &
    {\color{red}\xmark} \\
    
    \bottomrule
    \end{tabular}
    \vspace{2pt}

    \raggedright\footnotesize
    \hspace{2.1em} \textit{Notes.} \textbf{Parallel} denotes built-in vectorized environments for RL (not just GPU rendering). \textbf{Tactile Sensing} denotes built-in tactile-array observations with ready-to-use tactile end-effectors. \textbf{Hand Teleop} denotes built-in teleoperation of dexterous fingers, for example with data gloves or hand tracking.
\end{table*}

%% file: Dexsuite/imitation_results.tex
\begin{table*}[!t]
\caption{Baseline results on the DexSuite benchmark. Success rates (\%) are the mean $\pm$ standard deviation across three seeds with 50 episodes each. Track 1 tasks are completed with both the parallel-jaw gripper and the Allegro hand, and Track 2 tasks with a dexterous hand only, the Allegro hand or the Sharpa hand for \textit{Spray Graffiti} and \textit{Toast Bread}. Bold marks the best result for each task-embodiment pair, once among state-based and once among image-based policies, and results of 2\% or less are not bolded.}
\label{tab:il_results}
\centering
\scalebox{0.9}{%
\begin{minipage}{\textwidth}
\renewcommand{\arraystretch}{1.15}
\setlength{\tabcolsep}{3pt}
\begin{tabular*}{\textwidth}{@{\extracolsep{\fill}} l cccccccc @{}}
\toprule
& \multicolumn{5}{c}{\textit{Track 1}} & \multicolumn{3}{c}{\textit{Track 2}} \\
\cmidrule(lr){2-6}\cmidrule(l){7-9}
& \multicolumn{5}{c}{\textbf{Parallel-Jaw Gripper}} & \multicolumn{3}{c}{\textbf{Allegro Hand}} \\
\cmidrule(lr){2-6}\cmidrule(l){7-9}
\textbf{Model / Input} & \textbf{Reach} & \textbf{Lift} & \textbf{Push} & \textbf{Pick \& Place} & \textbf{Stack} & \textbf{Rack Mug} & \textbf{Move Pan} & \textbf{Drill to Point} \\
\midrule
BC-G (state) & $78 \textcolor{sdgray}{\,\pm\, 5}$ & $31 \textcolor{sdgray}{\,\pm\, 5}$ & $27 \textcolor{sdgray}{\,\pm\, 6}$ & $49 \textcolor{sdgray}{\,\pm\, 14}$ & $0 \textcolor{sdgray}{\,\pm\, 0}$ & $25 \textcolor{sdgray}{\,\pm\, 8}$ & $5 \textcolor{sdgray}{\,\pm\, 5}$ & $1 \textcolor{sdgray}{\,\pm\, 1}$ \\
BC-GMM (state) & $76 \textcolor{sdgray}{\,\pm\, 2}$ & $32 \textcolor{sdgray}{\,\pm\, 5}$ & $23 \textcolor{sdgray}{\,\pm\, 6}$ & $24 \textcolor{sdgray}{\,\pm\, 6}$ & $0 \textcolor{sdgray}{\,\pm\, 0}$ & $28 \textcolor{sdgray}{\,\pm\, 2}$ & $13 \textcolor{sdgray}{\,\pm\, 6}$ & $5 \textcolor{sdgray}{\,\pm\, 2}$ \\
BC-T (state) & $90\textcolor{sdgray}{\,\pm\, 2}$ & $57\textcolor{sdgray}{\,\pm\, 8}$ & $48\textcolor{sdgray}{\,\pm\, 4}$ & $64\textcolor{sdgray}{\,\pm\, 0}$ & $2\textcolor{sdgray}{\,\pm\, 0}$ & $58 \textcolor{sdgray}{\,\pm\, 3}$ & $68 \textcolor{sdgray}{\,\pm\, 16}$ & $47\textcolor{sdgray}{\,\pm\, 8}$ \\
Diffusion (state) & $\bm{100} \textcolor{sdgray}{\,\pm\, 0}$ & $\bm{72} \textcolor{sdgray}{\,\pm\, 3}$ & $\bm{63} \textcolor{sdgray}{\,\pm\, 2}$ & $\bm{65} \textcolor{sdgray}{\,\pm\, 5}$ & $\bm{16} \textcolor{sdgray}{\,\pm\, 2}$ & $\bm{82} \textcolor{sdgray}{\,\pm\, 2}$ & $\bm{69}\textcolor{sdgray}{\,\pm\, 6}$ & $\bm{50}\textcolor{sdgray}{\,\pm\, 2}$ \\
\midrule
ACT (image) & $76 \textcolor{sdgray}{\,\pm\, 6}$ & $\bm{73}\textcolor{sdgray}{\,\pm\, 6}$ & $\bm{50} \textcolor{sdgray}{\,\pm\, 2}$ & $\bm{66} \textcolor{sdgray}{\,\pm\, 4}$ & $1 \textcolor{sdgray}{\,\pm\, 1}$ & $37 \textcolor{sdgray}{\,\pm\, 6}$ & $\bm{89} \textcolor{sdgray}{\,\pm\, 3}$ & $\bm{43} \textcolor{sdgray}{\,\pm\, 3}$ \\
Diffusion (image) & $\bm{80} \textcolor{sdgray}{\,\pm\, 6}$ & $54 \textcolor{sdgray}{\,\pm\, 8}$ & $33 \textcolor{sdgray}{\,\pm\, 6}$ & $45 \textcolor{sdgray}{\,\pm\, 4}$ & $2 \textcolor{sdgray}{\,\pm\, 0}$ & $\bm{56} \textcolor{sdgray}{\,\pm\, 3}$ & $72\textcolor{sdgray}{\,\pm\, 14}$ & $32 \textcolor{sdgray}{\,\pm\, 4}$ \\
\bottomrule
\end{tabular*}

\par\medskip
\begin{tabular*}{\textwidth}{@{\extracolsep{\fill}} l cccccccc @{}}
\toprule
& \multicolumn{5}{c}{\textit{Track 1}} & \multicolumn{3}{c}{\textit{Track 2}} \\
\cmidrule(lr){2-6}\cmidrule(l){7-9}
& \multicolumn{6}{c}{\textbf{Allegro Hand}} & \multicolumn{2}{c}{\textbf{Sharpa Hand}} \\
\cmidrule(lr){2-7}\cmidrule(l){8-9}
\textbf{Model / Input} & \textbf{Reach} & \textbf{Lift} & \textbf{Push} & \textbf{Pick \& Place} & \textbf{Stack} & \textbf{Place Fruits} & \textbf{Spray Graffiti} & \textbf{Toast Bread} \\
\midrule
BC-G (state) & $82 \textcolor{sdgray}{\,\pm\, 9}$ & $21 \textcolor{sdgray}{\,\pm\, 2}$ & $46\textcolor{sdgray}{\,\pm\, 7}$ & $15 \textcolor{sdgray}{\,\pm\, 4}$ & $0 \textcolor{sdgray}{\,\pm\, 0}$ & $1 \textcolor{sdgray}{\,\pm\, 1}$ & $0 \textcolor{sdgray}{\,\pm\, 0}$ & $0 \textcolor{sdgray}{\,\pm\, 0}$ \\
BC-GMM (state) & $75\textcolor{sdgray}{\,\pm\, 4}$ & $43 \textcolor{sdgray}{\,\pm\, 8}$ & $47 \textcolor{sdgray}{\,\pm\, 5}$ & $35 \textcolor{sdgray}{\,\pm\, 5}$ & $0 \textcolor{sdgray}{\,\pm\, 0}$ & $3 \textcolor{sdgray}{\,\pm\, 1}$ & $0 \textcolor{sdgray}{\,\pm\, 0}$ & $0 \textcolor{sdgray}{\,\pm\, 0}$ \\
BC-T (state) & $88 \textcolor{sdgray}{\,\pm\, 2}$ & $47 \textcolor{sdgray}{\,\pm\, 6}$ & $60 \textcolor{sdgray}{\,\pm\, 7}$ & $\bm{76}\textcolor{sdgray}{\,\pm\, 5}$ & $1\textcolor{sdgray}{\,\pm\, 1}$ & $\bm{9} \textcolor{sdgray}{\,\pm\, 2}$ & $0 \textcolor{sdgray}{\,\pm\, 0}$ & $0 \textcolor{sdgray}{\,\pm\, 0}$ \\
Diffusion (state) & $\bm{97} \textcolor{sdgray}{\,\pm\, 1}$ & $\bm{65}\textcolor{sdgray}{\,\pm\, 8}$ & $\bm{68} \textcolor{sdgray}{\,\pm\, 5}$ & $67 \textcolor{sdgray}{\,\pm\, 3}$ & $\bm{9} \textcolor{sdgray}{\,\pm\, 5}$ & $7 \textcolor{sdgray}{\,\pm\, 3}$ & $0.7 \textcolor{sdgray}{\,\pm\, 1}$ & $0 \textcolor{sdgray}{\,\pm\, 0}$ \\
\midrule
ACT (image) & $\bm{73} \textcolor{sdgray}{\,\pm\, 1}$ & $45 \textcolor{sdgray}{\,\pm\, 5}$ & $\bm{66} \textcolor{sdgray}{\,\pm\, 6}$ & $\bm{60} \textcolor{sdgray}{\,\pm\, 7}$ & $0 \textcolor{sdgray}{\,\pm\, 0}$ & $\bm{27} \textcolor{sdgray}{\,\pm\, 5}$ & $0 \textcolor{sdgray}{\,\pm\, 0}$ & $0 \textcolor{sdgray}{\,\pm\, 0}$ \\
Diffusion (image) & $61 \textcolor{sdgray}{\,\pm\, 2}$ & $\bm{58} \textcolor{sdgray}{\,\pm\, 12}$ & $62\textcolor{sdgray}{\,\pm\, 3}$ & $52 \textcolor{sdgray}{\,\pm\, 11}$ & $1 \textcolor{sdgray}{\,\pm\, 1}$ & $5 \textcolor{sdgray}{\,\pm\, 1}$ & $0.7 \textcolor{sdgray}{\,\pm\, 1}$ & $0 \textcolor{sdgray}{\,\pm\, 0}$ \\
\bottomrule
\end{tabular*}
\end{minipage}%
}
\end{table*}